\documentclass{article}
\usepackage{spconf,graphicx}
\usepackage[fleqn]{amsmath}
\usepackage{amsfonts}
\usepackage{amssymb}
\usepackage{url}
\usepackage{float}
\usepackage{dsfont}
\usepackage{algorithm}
\usepackage{algpseudocode}
\usepackage{color}

\usepackage[table,xcdraw]{xcolor}
\DeclareMathOperator*{\argmax}{arg\,max}

\newcommand{\var}[1]{{\ttfamily#1}}

\title{MULTI-SCALE ALIGNMENT AND CONTEXTUAL HISTORY FOR ATTENTION
	MECHANISM IN SEQUENCE-TO-SEQUENCE MODEL}
\name{Andros Tjandra\textsuperscript{1,2}, Sakriani Sakti\textsuperscript{1,2}, Satoshi Nakamura\textsuperscript{1,2}}
\address{\textsuperscript{1} Graduate School of Information Science, Nara Institute of Science and Technology, Japan\\\textsuperscript{2} RIKEN, Center for Advanced Intelligence Project AIP, Japan\\
	\texttt{\{andros.tjandra.ai6, ssakti, s-nakamura\}@is.naist.jp}}

\begin{document}
%
\maketitle
\begin{abstract}
A sequence-to-sequence model is a neural network module for mapping two sequences of different lengths. The sequence-to-sequence model has three core modules: encoder, decoder, and attention. Attention is the bridge that connects the encoder and decoder modules and improves model performance in many tasks. In this paper, we propose two ideas to improve sequence-to-sequence model performance by enhancing the attention module. First, we maintain the history of the location and the expected context from several previous time-steps. Second, we apply multiscale convolution from several previous attention vectors to the current decoder state. We utilized our proposed framework for sequence-to-sequence speech recognition and text-to-speech systems. The results reveal that our proposed extension could improve performance significantly compared to a standard attention baseline.
\end{abstract}
\begin{keywords}
sequence-to-sequence model, attention mechanism, multiscale alignment, contextual history, ASR and TTS
\end{keywords}

\vspace{-0.1cm}
\section{Introduction}
\label{sec:intro}
\vspace{-0.15cm}
The deep neural network has become a popular machine learning module for solving many different tasks due to its flexibility and high performance. In the case of modeling the transformation from a dynamic-length input into a dynamic-length output, a specific neural network architecture called the sequence-to-sequence model has been proposed \cite{sutskever2014sequence}. The first sequence-to-sequence model used two main modules, the encoder and decoder. The encoder module captures the overall information from the input source, and the decoder module generates the output target.

Training the encoder-decoder model is difficult because the decoder depends on the last hidden layer from the encoder state. The decoding results become degraded when the source input or target output is too long. Therefore, an attention mechanism \cite{bahdanau2014neural} was proposed to overcome the encoder-decoder limitations by introducing a ``shortcut'' connection between the decoder and encoder states. The attention mechanism allows the model to focus on a certain part from the encoder and neglects the other information. Following the success of this attention mechanism, most attention-based models have been applied to many types of variable-length transformation tasks (e.g., machine translation (text-to-text systems) \cite{bahdanau2014neural, sutskever2014sequence} and speech recognition (speech-to-text systems) \cite{chan2016listen, chorowski2015attention}). In extreme cases, a model named Transformers \cite{vaswani2017attention}, built based on a stack of feedforward layer in addition to self-attention, is able to improve the NMT results significantly.

Considering the importance of attention mechanism, various research works have been conducted on modifying the attention module. In most cases, the alignment from the attention vector is calculated from the entire input source sequence and normalized by a softmax function. One work \cite{luong2015effective} proposed the approach of calculating only a subset of input sequences, thus reducing the computational cost for each decoding step. Extending local attention, other works \cite{tjandra17local, raffel2017online} proposed a new mechanism to enforce monotonicity and enable online decoding with the sequence-to-sequence model. Some research \cite{xu2015show} proposed replacing soft attention with a hard attention mechanism by sampling categorical random variables and training through maximizing the variational lower bound from across multiple samples. Another hard-attention mechanism has been utilized \cite{do2017toward} to extract the emphasis level for speech-to-speech translation.

Based on the general attention module, some papers have proposed different scoring systems to calculate the ``relevance'' score between the encoder and decoder states. However, most of these works did not incorporate information other than the encoder state and the current decoder state. To date, only the work by Chorowski et al. attempted to integrate knowledge of the previous alignment probability in the scoring function \cite{chorowski2015attention}. Nevertheless, almost no works that have attempted to incorporate information of the prior time-step to assist in the scoring function.

In contrast, in this work, we propose a novel attention-scoring mechanism that integrates: (1) the alignment with multiscale convolution, and (2) the contextual information from the history of multiple time-steps. We utilized our proposed framework for sequence-to-sequence speech recognition and text-to-speech systems. From our experiment, it is proved that each extension is able to improve the ASR and TTS performance and robustness.

We first introduce the sequence-to-sequence model and the importance of the attention module for achieving better performance in Section~\ref{sec:intro}. Sequence-to-sequence ASR and TTS based on encoder-decoder architecture is described in Section~\ref{sec:encdecasr}. Then, Section~\ref{sec:proposed} provides the formulation details of our proposed method. Section~\ref{sec:exp_asr} and \ref{sec:exp_tts} present our experimental set-up (including the dataset, feature extraction, precise model architecture), and experiment results of our baseline and proposed attention module for ASR and TTS, respectively. Finally, we finish our paper with our conclusions and possible future extension of this work in Section~\ref{sec:concl}.

\vspace{-0.1cm}
\section{Sequence-to-Sequence Architecture}
\label{sec:encdecasr}
\vspace{-0.15cm}
The sequence-to-sequence model is a type of neural network model that directly models conditional probability $P(\mathbf{y}|\mathbf{x})$, where $\mathbf{x} = [x_1, ..., x_S]$ is the source sequence with length $S$, and $\mathbf{y} = [y_1, ..., y_T]$ is the target sequence with length $T$.

The overall structure of the attention-based encoder-decoder model consists of three modules: encoder, decoder, and attention modules. The encoder task processes input sequence $\mathbf{x}$ and outputs representative information $\mathbf{h^E} = [h^E_1, ...,h^E_S] \in \mathbb{R}^{S \times N}$ for the decoder. The attention module is an extension scheme that helps the decoder find relevant information on the encoder side based on the current decoder's hidden states \cite{bahdanau2014neural}. An attention module produces context information $c_t$ at time $t$ based on the encoder's and decoder's hidden states with the following equation:
\begin{align}
c_t &= \sum_{s=1}^{S} a_t[s] * h^E_s \label{eq:exp_ctx}\\
a_t[s] &= \text{Align}({h^E_s}, h^D_t) \\&= \frac{\exp(\text{Score}(h^E_s, h^D_t))}{\sum_{s=1}^{S}\exp(\text{Score}(h^E_s, h^D_t))} \label{eq:align},
\end{align}
where $\text{Score:}(\mathbb{R}^M \times \mathbb{R}^N) \rightarrow \mathbb{R}$, $M$ is the number of hidden units of the encoder and $N$ is the number of hidden units of the decoder.
Finally, the decoder task, which predicts the target sequence probability at time $t$ based on the previous output and context information $c_t$, can be formulated:
\vspace{-0.2cm}
\begin{equation}
\log{P(\mathbf{y}|\mathbf{x}; \theta)} = \sum_{t=1}^{T}\log{P(y_t|h_t^D, c_t; \theta)} ,\label{eq:mle}
\end{equation} where $h_t^D$ is the last decoder layer that contains summarized information from all previous inputs $\mathbf{y}_{<t}$ and $\theta$ is our model parameter.

\vspace{-0.1cm}
\subsection{Attention Module}
\vspace{-0.2cm}
The scoring function $\text{Score}$ is used to calculate the ``relevance'' value between a source side and a target side. The following are some of the most common forms for function $\text{Score}$:
\begin{itemize}
	\item Dot product:\\
	\vspace{-0.2cm}
	\begin{equation}
	Score(h_s^E, h_t^D) = \sum_{m=1}^{M}{h_s^E[m] * h_t^D[m]}, \end{equation}

	where $h_s^E, h_t^D \in \mathbb{R}^{M}$ and there is no trainable parameter.
	\item Bilinear product:\\
	\vspace{-0.2cm}
	\begin{equation}
	Score(h_s^E, h_t^D) = h_s^E W h_t^D,
	\end{equation}
	where $h_s^E \in \mathbb{R}^{M}, h_t^D \in \mathbb{R}^{N}$ and there is a trainable parameter $W \in \mathbb{R}^{M \times N}$.
	\item MLP attention:\\
	\vspace{-0.2cm}
	\begin{equation}
	Score(h_s^E, h_t^D) = W_3 \, \tanh(W_1 h_s^E + W_2 h_t^D), \label{eq:mlp_att}
	\end{equation} 
	where $h_s^E \in \mathbb{R}^{M}, h_t^D \in \mathbb{R}^{N}$ and there are trainable parameters $W_1 \in \mathbb{R}^{P \times M}, W_2 \in \mathbb{R}^{P \times N}, W_3 \in \mathbb{R}^{1 \times P}$.
\end{itemize}
The result from function $\text{Score}$ is an unnormalized scalar value. If the value is positive, then the scorer module indicates that the encoder state at time-step $s$ is relevant with the current decoder state, but if the value is negative, the scorer module indicates that the encoder state at time-step $s$ is irrelevant. Therefore, the score result will be normalized across all encoder states $h^E = [h_1^E,..,h_S^E]$ with the $\text{softmax}$ function.

\vspace{-0.1cm}
\subsection{Sequence-to-Sequence ASR}
\begin{figure}[]
	\centering
	\includegraphics[width=0.75\linewidth]{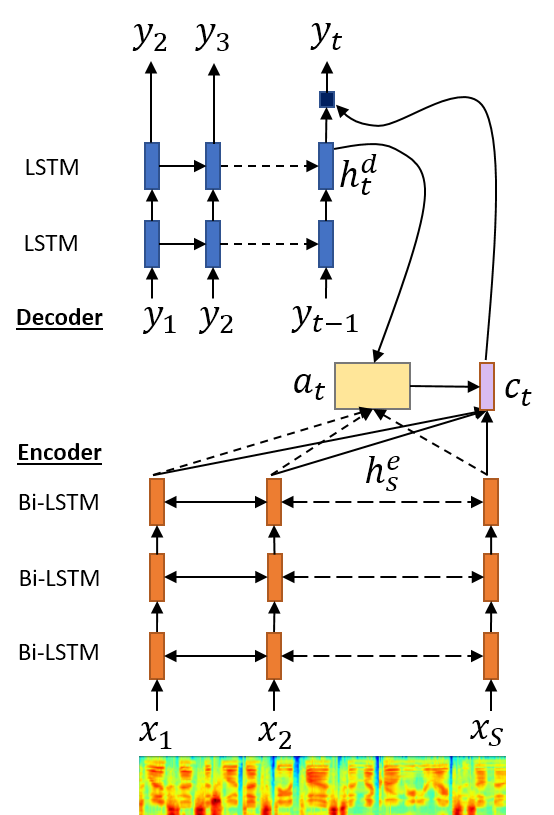}
	\caption{Attention-based encoder-decoder ASR.}
	\label{fig:atte2e_asr}
\end{figure}

\vspace{-0.2cm}
For a sequence-to-sequence model applied to the ASR task, $\mathbf{x}$ is a sequence of feature vectors like Mel-spectral filterbank and/or MFCC. Therefore, $\mathbf{x} \in \mathbb{R}^{S \times F}$, where F is the number of features and S is the total frame length for an utterance. Output $\mathbf{y}$, which is a speech transcription sequence, can be either a phoneme or a grapheme (character) sequence. We illustrate an attention-based ASR in Figure \ref{fig:atte2e_asr}.

\vspace{-0.1cm}
\subsection{Sequence-to-Sequence TTS} \label{sec:seq2seq_tts}
\begin{figure}[]
\centering
\includegraphics[width=0.70\linewidth]{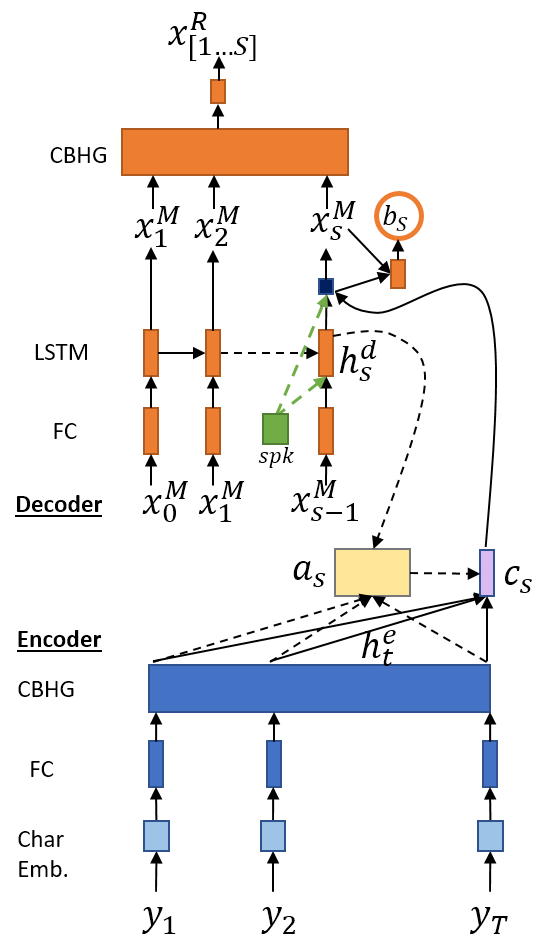}
\caption{Attention-based encoder-decoder TTS.}
\label{fig:atte2e_tts}
\end{figure}

\vspace{-0.2cm}
For a sequence-to-sequence module applied to the TTS task (see Figure~\ref{fig:atte2e_tts}), $\mathbf{y}$ is a character sequence, $\mathbf{x}^M$ is a mel-spectrogram frame, and $\mathbf{x}^R$ is a linear-spectrogram frame. Both $\mathbf{x}^{M}$ and $\mathbf{x}^R$ have same sequence length $S$.

In this work, we use Tacotron \cite{wang2017tacotron} model with binary frame-ending prediction \cite{tjandra2017listening, tjandra2018machine}. In the training stage, we optimized the model by minimizing the following loss function:
\vspace{-0.3cm}
	\begin{equation}
	\vspace{-0.2cm}
	\begin{split}
	Loss_{TTS}(x, \hat{x}, b, \hat{b}) &= \sum_{s=1}^{S} ||x_s^M - \hat{x}_s^M||^2 + ||x_s^R - \hat{x}_s^R||^2 \\
	& - (b_s \log(\hat{b}_s) + (1-b_s) \log(1-\hat{b}_s))
	\end{split}
	\end{equation}
where $\hat{x}^M_S, \hat{x}^R_s, \hat{b}_s$ are the predicted log-Mel spectrogram, the log linear spectrogram and the end of frame probability, and $x^M_s, x^R_s, b_s$ is the ground truth.

\section{Incorporating Multiscale Alignment and Contextual History} 
\label{sec:proposed}
This section present our proposed attention-scoring mechanism that integrates: (1) the alignment with multiscale convolution, and (2) the contextual information from the history of multiple time-steps. 

For each time-step $t$, the decoder state $h_t^D$ generate an alignment vector $a_t \in \mathbb{R}^{S}$ and an expected context $c_t \in \mathbb{R}^{M}$. In most of attention encoder-decoder module, $c_t$ is the input for top layer like softmax layer. Here, instead of only used once, we keep the past information up to $o$-order time-step $a_{t-o},a_{t-o+1}...,a_{t-1}$ and $c_{t-o}, c_{t-o+1}, ...,c_{t-1}$ in a queue data structures. 

\begin{figure*}[]
	\centering
	\includegraphics[width=0.8\linewidth]{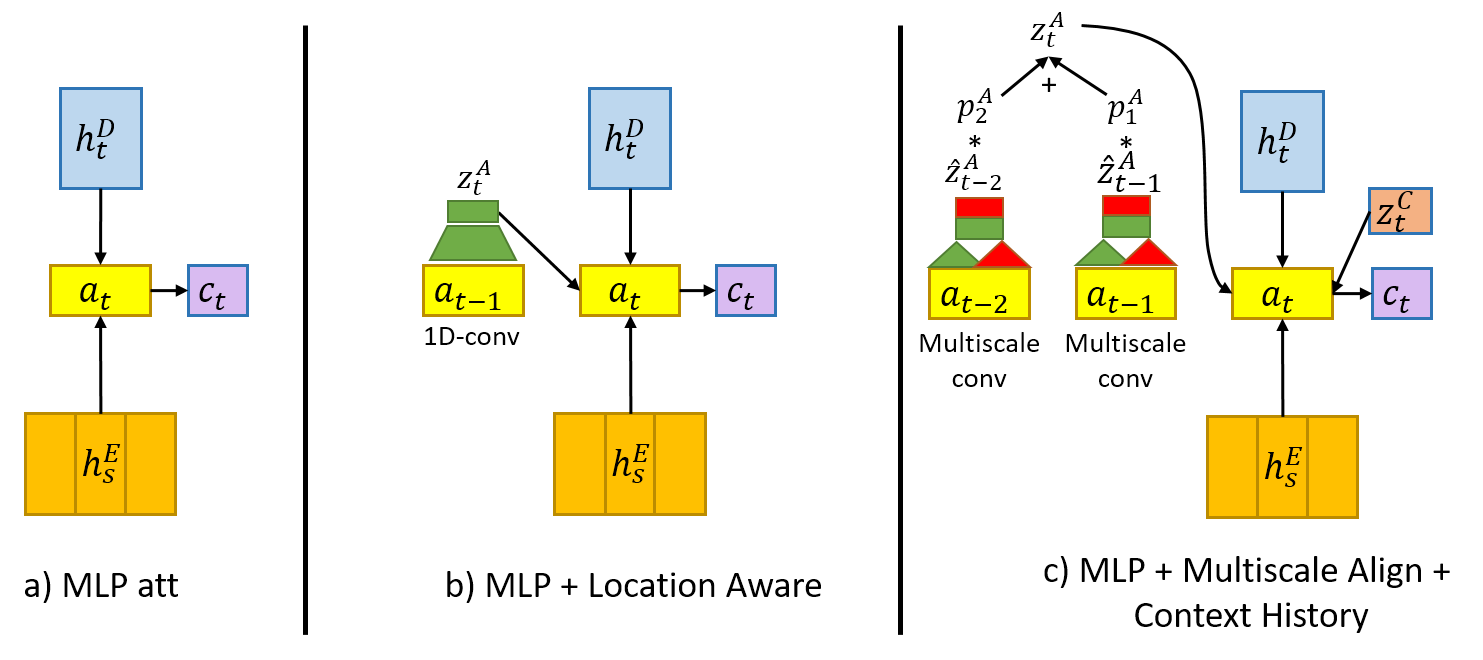}
	\caption{Comparison between several attention mechanisms: (a) MLP, (b) MLP + Location Aware, and (c) our proposed method, side by side}
    \vspace{-0.2cm}
	\label{fig:compare}
\end{figure*}

Figure~\ref{fig:compare} illustrates a comparison between several attention mechanisms: (a) MLP, (b) MLP + Location Aware \cite{chorowski2015attention}, and (c) our proposed method. The details of our proposed approach is described below. 

\subsection{Multiscale Alignment Information}
\begin{figure}[htb]
	\centering
	\includegraphics[width=0.90\linewidth]{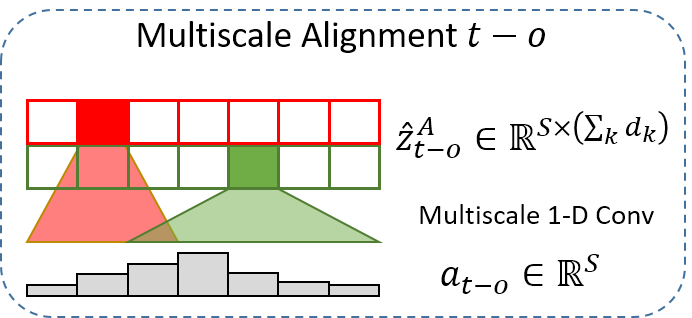}
	\caption{Multiscale alignment for alignment probability at time $t-o$}
	\label{fig:mscale}
\end{figure}
In order to extract information from speech of different lengths, we apply multiscale 1-D convolution to the previous attention vector. Compared to a single 1-D convolution, we could capture different information depending on the kernel size. Here, the input is a normalized alignment probability from $o$ previous time-step $a_{t-o} = [a_{t-o}[1],..,a_{t-o}[S]] \in \mathbb{R}^{S}$. We have multiple $K$ different sets for the 1-D filter $\mathcal{F} = {\mathbf{F_1, F_2, .., F_K}}$. For each $\mathbf{F_k} \in \mathbb{R}^{1 \times t_k \times d_k}$, there is a convolution filter with $t_k$ kernel size, $1$ input channel, and $d_k$ output channel. 

We apply convolution to $a_{t-o}$ across all filters $\mathcal{F}$ and concatenate the results based on the last axis (output channel):
\begin{align}
\hat{z}^A_{t-o} = f([\mathbf{F_1}*a_{t-o},...,\mathbf{F_k}*a_{t-o}]) \in \mathbb{R}^{S\times(\sum_{k}{d_k})} \label{eq:mscale_single}
\end{align} where $*$ is the ``same'' convolution operator that applies padding to keep the output length the same as the input $a_{t-o}$ and $f(\cdot)$ is a nonlinear activation function. In Figure~\ref{fig:mscale}, we visualize the multiscale convolution 1-D on the top of alignment $a_{t-o}$.

We apply Eq.~\ref{eq:mscale_single} across all $a_{t-o},.., a_{t-1}$ by sharing the filter $\mathcal{F}$ across multiple time-steps. Next, we merge the information:
\begin{equation}
z^A_t = \sum_{i=1}^{o} p^A_i * \hat{z}^A_{t-i} \label{eq:mscale_multi},
\end{equation} where $p_i^A \in [0..1]$ and  $\sum_{i=1}^{o} p_i^A = 1$. In this work, we treat $p_i^A$ as learnable parameters.

\subsection{Contextual History Information}
\begin{figure}[htb]
	\vspace{-0.3cm}
	\centering
	\includegraphics[width=0.50\linewidth]{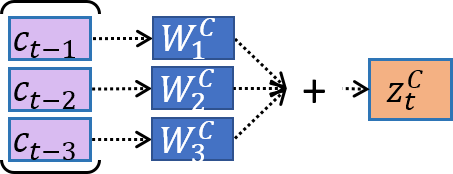}
	\caption{Integrating information from context history of multiple time-steps}
	\label{fig:ctxhist}
\end{figure}
To summarize a dynamic-length encoder state into a vector with fixed dimension $c_t$, the alignment probability $a_t$ is multiplied by $\mathbf{h^E}$ (see Eq.~\ref{eq:exp_ctx}). Providing information from the previous $c_{t-o}, c_{t-o+1}, ...,c_{t-1}$ is a useful way for the model to keep track of which information was under focus in the last $o$ time-step. To integrate the past contextual information $c_{t-o}, c_{t-o+1}, ..., c_{t-1}$ for the current decoder at time-step $t$, we calculate $z^C_t$ with
\begin{equation}
z^C_t = f\left(\sum_{i=1}^{o} \left(W_{i}^C c_{t-i} + b_{i}^C\right)\right) \in \mathbb{R}^{P_{ctx}} \label{eq:ctxhist},
\end{equation} where $W_{i}^C \in \mathbb{R}^{P_{ctx} \times M}, b_i^C\in \mathbb{R}^{P_{ctx}}$ and $f(\cdot)$ is a nonlinear activation function. In Figure~\ref{fig:ctxhist}, we visualize how to integrate the past contextual information together into a vector $z^C_t$.

\begin{algorithm}[h]
	\caption{An example of decoding algorithm for our proposed method}
    \label{alg:decode}
	\begin{algorithmic}[1]
		\State \textbf{Parameters} \var{OT}: time-step history
		\Procedure{Decode}{$\mathbf{h^E} \in \mathbb{R}^{S \times M}$: enc state}
		\State $qA$ = \var{Queue()} \Comment{queue for alignment history}
		\State $qC$ = \var{Queue()} \Comment{queue for context history}

		\State $y$ = [\var{<bos>}]; $t = 0$;
		\For{\var{i} in [0..\var{OT}]} \Comment{initialize queue}
			\State $qA$.\var{push}($[1,0,0,..,0] \in \mathbb{R}^S$)
			\State $qC$.\var{push}($[0,0,0,..,0] \in \mathbb{R}^M$)
		\EndFor

		\While{$y[t]$ $\neq$ \var{<eos>}} \Comment{eos = end-of-sentence}
		\State $z^A_t$= \var{MultiscaleConv1D($qA$)} \Comment{Eq.~\ref{eq:mscale_single}, \ref{eq:mscale_multi}}
		\State $z^C_t$ = \var{ContextHist($qC$)} \Comment{Eq.~\ref{eq:ctxhist}}
		\State $h_t^D$ = \var{LSTM($y[t], h_{t-1}^D$)}
		\For{\var{s} in [0..\var{S}]}
		\State $a_{t}[s]$ = \var{Align($h_s^E, h_t^D, z^A_t, z^C_t$)} \Comment{Eq.~\ref{eq:align},\ref{eq:score_combine}}
		\EndFor
		\State $c_{t}$ = $\sum_s a_t[s] * h^E_s$ \Comment{Eq.~\ref{eq:exp_ctx}}
		\State $qA$.\var{pop()}; \quad $qC$.\var{pop()};
		\State $qA$.\var{push}($a_{t}$); \quad $qC$.\var{push}($c_{t}$);
		\State $PY$ = \var{Softmax}($W c_{t} +b$); 		\State $t$ = $t+ 1$
		\State $y[t]$ = $\argmax_{c}$$PY[c]$ \Comment{greedy decoding}
		\EndWhile
		\State \var{return} $y$
		\EndProcedure
	\end{algorithmic}
\end{algorithm}

\subsection{MLP attention with Alignment History and Contextual Information}
To integrate all information from multiscale alignment history and contextual information into an MLP attention formula, we rewrite Eq.~\ref{eq:mlp_att} with the following equation:
\begin{align}
Score(h_s^E, h_t^D, &z^A_t, z^C_t) = W_5 \,\, g(W_1 h_s^E+ W_2 h_t^D \nonumber \\
&+ W_3 z^A_t[s] + W_4 z^C_t + b) \label{eq:score_combine}
\end{align} where $W_1 \in \mathbb{R}^{P_{info} \times M}$, $W_2 \in \mathbb{R}^{P_{info} \times N}$, \\$W_3 \in \mathbb{R}^{ P_{info} \times (\sum_k d_k)}$, $W_4 \in \mathbb{R}^{P_{info} \times P_{ctx}}$, $b \in \mathbb{R}^{P_{info}}$, $W_5 \in \mathbb{R}^{1\times P_{info}}$ and $g(\cdot)$ is a nonlinear activation function.
The details of algorithm can be seen at Alg.~\ref{alg:decode}. 

\section{Experiment for End-to-End ASR}
\label{sec:exp_asr}
\subsection{Dataset and Feature Representation}
In this study, we investigated the performance of our proposed method on WSJ \cite{paul92wsj} with identical definitions of training, development, and test sets to those of the Kaldi s5 recipe \cite{povey11asru}. We separated WSJ into two experiments using WSJ-SI84 only and WSJ-SI284 data for training. Our validation set was dev\_93 and our test set was eval\_92. 

We used the character sequence as our decoder target and followed the preprocessing steps proposed by \cite{hannun2014first}. The text from all utterances was mapped into a 32-character set: 26 (a-z) letters of the alphabet, apostrophes, periods, dashes, space, noise, and ``eos.'' In all experiments, we extracted the 40 dims + $\Delta$ + $\Delta\Delta$ (total 120 dimensions) log Mel-spectrogram features from our speech and normalized every dimension into zero mean and unit variance.

\subsection{Model Architecture}
\vspace{-0.2cm}
On the encoder side, we fed our input features into a linear layer with 512 hidden units followed by the LeakyReLU \cite{xu2015empirical} activation function. We used three bidirectional LSTMs (Bi-LSTM) for our encoder with 256 hidden units for each LSTM (total 512 hidden units for Bi-LSTM). To improve running time and reduce memory consumption, we used hierarchical subsampling \cite{graves2012supervised, bahdanau2016end} on the top three Bi-LSTM layers and reduced the number of encoder time-steps by a factor of 8.

On the decoder side, we used a 128-dimensional embedding matrix to transform the input graphemes into a continuous vector, followed by unidirectional LSTMs with 512 hidden units.

In the training phase, we optimized our model parameters based on the maximum likelihood estimation (MLE) criterion (Eq.~\ref{eq:mle}) using the sum along the time-step axis on the target side and divided by the number of sequences per mini-batch. To optimize our model, we used the Adam \cite{kingma2014adam} optimizer.

In the decoding phase, the transcription was generated via beam-search with beam size 5, and the beam score are normalized the log-likelihood, divided by the hypothesis length to prevent the decoder from favoring the shorter transcriptions. We did not use any language model or lexicon dictionary for all experiments.

\vspace{-0.1cm}
\subsection{Results}
\vspace{-0.2cm}
In this section, we provide experimental results and discuss how each piece of additional information is used.
\vspace{-0.3cm}
\begin{table}[]
	\centering

	\begin{tabular}{|l|c|}
		
		\hline
		\multicolumn{1}{|c|}{\textbf{Model}}                    & \textbf{CER (\%)} \\ \hline
		\multicolumn{2}{|c|}{\cellcolor[HTML]{EFEFEF}\textbf{Baseline}}             \\ \hline
		MLP \cite{kim2017joint} & 11.08 \\ \hline
		MLP+Location Aware \cite{kim2017joint} & 8.17              \\ \hline
		MLP \cite{tjandra2017attention} (ours)                                              & 7.12              \\ \hline
		MLP+Location Aware (ours, t=15)                         & 6.87               \\ \hline
		\multicolumn{2}{|c|}{\cellcolor[HTML]{EFEFEF}\textbf{Proposed}}             \\ \hline
		MLP + MA (O=1)                                          & 6.43              \\ \hline
		MLP + MA + C (O=1)                                      & 6.04              \\ \hline
		MLP + MA + C (O=2)                                      & 5.85              \\ \hline
		MLP + MA + C (O=3)                                      & 5.59              \\ \hline
	\end{tabular}
	\caption{MA denotes multiscale aligment, C denotes contextual history, and O denotes how many time-steps are incorporated in the history for the scoring function.} 	\label{tbl:asr_res}
\end{table}

\vspace{-0.1cm}
\subsubsection{Baseline MLP attention}
\vspace{-0.1cm}
For our baseline model, we used a standard MLP attention module \cite{bahdanau2014neural} (Eq.~\ref{eq:mlp_att}) and a standard MLP + location-aware attention module \cite{bahdanau2016end}. In both attention modules, we have a projection layer with 256 hidden units. For the MLP + location-aware attention module, we used a 1-D convolutional filter with kernel size 15.

\vspace{-0.1cm}
\subsubsection{MLP Attention +  Multiscale Alignment + Context History on Multiple Timestep History}
\vspace{-0.1cm}
In this section, we add the multiscale 1-D convolution filter and context history to our baseline results. Here, we used a set of 1-D convolution filters $\mathcal{F}$ with output channel $d_k=64$ and kernel size $t_1 = 7, t_2 = 15, t_3=31, t_4=63$. For the activation $f(\cdot)$ (Eq.~\ref{eq:mscale_multi}, \ref{eq:ctxhist}) is LeakyReLU function\cite{xu2015empirical} and $g(\cdot)$ (Eq.~\ref{eq:score_combine}) is tanh function. From Table
\ref{tbl:asr_res}, we can see that by combining multiscale alignment and context history, we get significant improvement compared to the normal MLP + location-aware module. Furthermore, by increasing the number of time-steps incorporated in the history, we get yet further improvement. Our best model with multiscale alignment and contextual information from the last three time-steps achieved 5.59\% CER.

\vspace{-0.1cm}
\section{Experiment for End-to-End TTS}
\vspace{-0.2cm}
\label{sec:exp_tts}
\begin{figure*}[t]
	\vspace{-0.3cm}
	\centering
	\includegraphics[width=1.0\linewidth]{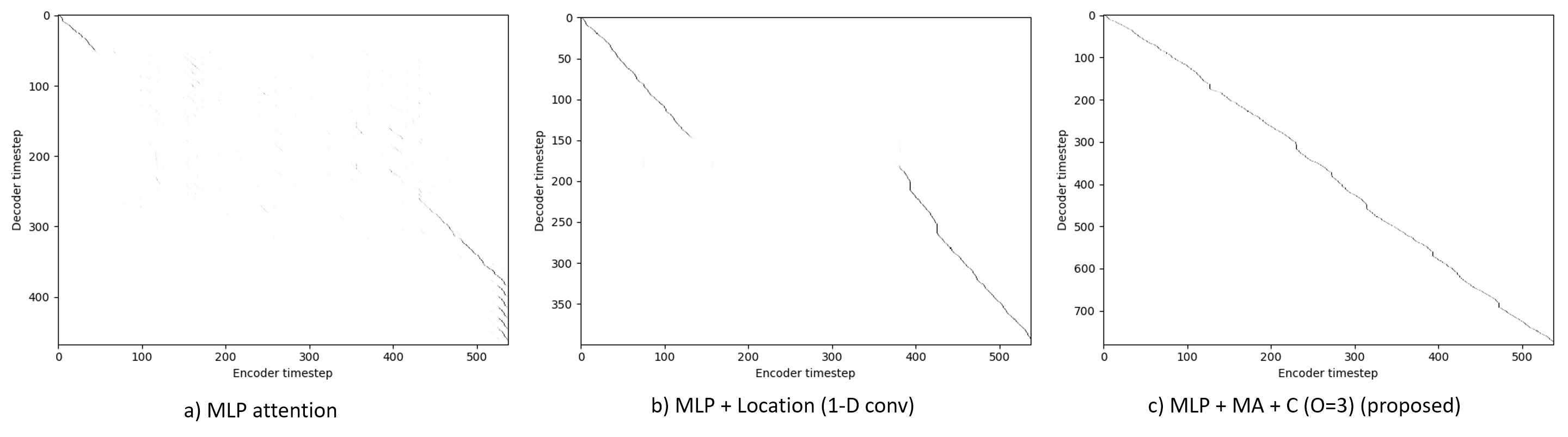}
	\caption{Attention plot from a very long text with a) MLP attention, b) MLP + Location aware (1-D conv), c) MLP + Multiscale Alignment + Contextual History with time-step order=3}
	\label{fig:plot_tts_all}
\end{figure*}

\vspace{-0.1cm}
\subsection{Dataset and Feature Representation}
\vspace{-0.2cm}
We investigated the performance of Tacotron with different attention module on LJSpeech \cite{ljspeech17} dataset. Because LJSpeech doesn't have official train, development and test set, we split the dataset by 94\% for training (12314 utts), 3\% for development (393 utts) and 3\% for testing (393 utts). For the feature extraction, we extracted the log-linear spectrogram with 50-ms window length, 12.5-ms step size, and 2048-point short-time Fourier transform (STFT) \cite{mcfee2015librosa}. Then, we also extracted the 80-dimensional log-Mel spectrogram. Both log-linear and log-Mel spectrogram are used as the target for Tacotron model. To represent the text, we tokenized and feed the Tacotron encoder in the character level representation.

\vspace{-0.1cm}
\subsection{Model Architecture}
\vspace{-0.2cm}
We use Tacotron with binary-frame ending prediction \ref{sec:seq2seq_tts}. We left the hyperparameters same as those for the original Tacotron, except we replaced the activation function ReLU with the LReLU function. For the CBHG module, we used $K=8$ filter banks instead of 16 to reduce the GPU memory consumption. For the decoder sides, we deployed two LSTMs instead of GRU with 256 hidden units. For each time-step, our model generated 4 consecutive frames to reduce the number of steps in the decoding process.

\vspace{-0.1cm}
\subsection{Result}
\vspace{-0.2cm}
\begin{table}[]
	\centering
	\begin{tabular}{|l|c|}
		\hline
		\multicolumn{1}{|c|}{\textbf{Model}}     & \textbf{L2-norm}     \\ \hline
		\multicolumn{2}{|c|}{\cellcolor[HTML]{EFEFEF}\textbf{Baseline}} \\ \hline
		MLP                                      & 0.653                \\ \hline
		MLP + Location Aware                     & 0.654                \\ \hline
		\multicolumn{2}{|c|}{\cellcolor[HTML]{EFEFEF}\textbf{Proposed}} \\ \hline
		MLP + MA + C (O=1)                       & 0.644                \\ \hline
		MLP + MA + C (O=2)                       & 0.636                \\ \hline
		MLP + MA + C (O=3)                       & 0.629                \\ \hline
	\end{tabular}
	\caption{MA denotes multiscale aligment, C denotes contextual history, and O denotes how many time-steps are incorporated in the history for the scoring function.} 	\label{tbl:tts_res}
\end{table}

In Table~\ref{tbl:tts_res}, we shows the experiment results from baseline MLP, MLP + Location Aware and our proposed method on LJSpeech test set. Here, we measure the L2-norm squared between the ground-truth and the predicted log-Mel spectrogram.

For the further analysis, we also investigate TTS to generate speech from a very long unseen text from WSJ1 (total 537 characters): \\

\noindent \fbox{\begin{minipage}{24em}
\textit{``he says that he doesn't believe drexel customers' holdings in staley were ever large enough to require a thirteen d. filing., following the wilson affair the panel did conduct several staff inquiries in reaction to news stories, one yielded reprimands of representatives, daniel crane r. illinois and gerry studds d. massachusetts in nineteen eighty three for sexual affairs with house pages, representative george hansen r. idaho drew a reprimand in nineteen eighty four after a felony conviction for falsifying his financial disclosures.''}
\end{minipage}}\\

The attention matrix is plotted in Fig.~\ref{fig:plot_tts_all}(c). For comparison, we also plotted the attention matrix from the results of the attention baseline (see Fig.~\ref{fig:plot_tts_all}(a) and (b)). As can be seen, from three different attention module, only our proposed model are able to synthesize the log-Mel spectrogram from the start until the end. Our TTS samples could be found at \url{https://ttssample2018v1.netlify.com}.

\vspace{-0.2cm}
\section{Conclusion}
\label{sec:concl}
\vspace{-0.2cm}
In this work, we proposed a new attention-scoring mechanism by integrating the alignment and the contextual information from the history of multiple time-steps. The information from previous results was added in order to assist in the scoring function used to calculate the relevance information between the encoder state and the decoder state. From our experiment, the proposed scoring function achieved significant improvement over the baseline result in both ASR and TTS experiment. In future work, we intend to integrate a more sophisticated method, such as a differentiable memory network, to replace the queue mechanism.

\vspace{-0.2cm}
\section{Acknowledgement}
\label{sec:ack}
\vspace{-0.2cm}
Part of this work was supported by JSPS KAKENHI Grant Numbers JP17H06101 and JP17K00237.

\bibliographystyle{IEEEbib}
\bibliography{strings,refs}

\end{document}